\documentclass[10pt]{cai}

\begin{document}
\def\conferenceyear{2025}
\volumeheader{38}{0}
\begin{center}


\title{InceptoFormer: A Multi-Signal Neural Framework for Parkinson’s Disease Severity Evaluation from Gait}
\maketitle

\thispagestyle{empty}

\begin{tabular}{cc}
Safwen Naimi\upstairs{\affilone,*},
Arij Said\upstairs{\affilone},
Wassim Bouachir\upstairs{\affilone}, Guillaume-Alexandre Bilodeau\upstairs{\affiltwo}
\\[0.25ex]
{\small \upstairs{\affilone} University of Quebec (TÉLUQ), Montreal, QC, Canada} \\
{\small \upstairs{\affiltwo} Polytechnique Montréal, Montreal, QC, Canada} \\
 \\
\end{tabular}
  
\emails{
  \upstairs{*}safwen.naimi@teluq.ca 
}
\vspace*{0.2in}
\end{center}

\begin{abstract}
We present \textit{InceptoFormer}, a multi-signal neural framework designed for Parkinson’s Disease (PD) severity evaluation via gait dynamics analysis. Our architecture introduces a 1D adaptation of the Inception model, which we refer to as Inception1D, along with a Transformer-based framework to stage PD severity according to the Hoehn and Yahr (H\&Y) scale. The Inception1D component captures multi-scale temporal features by employing parallel 1D convolutional filters with varying kernel sizes, thereby extracting features across multiple temporal scales. The transformer component efficiently models long-range dependencies within gait sequences, providing a comprehensive understanding of both local and global patterns. To address the issue of class imbalance in PD severity staging, we propose a data structuring and preprocessing strategy based on oversampling to enhance the representation of underrepresented severity levels.
The overall design enables to capture fine-grained temporal variations and global dynamics in gait signal, significantly improving classification performance for PD severity evaluation.
Through extensive experimentation, \textit{InceptoFormer} achieves an accuracy of 96.6\%, outperforming existing state-of-the-art methods in PD severity assessment. The source code for our implementation is publicly available at \href{https://github.com/SafwenNaimi/InceptoFormer-A-Multi-Signal-neural-framework-for-Parkinson-s-Disease-Severity-Evaluation-from-Gait}{https://github.com/SafwenNaimi/InceptoFormer}. 
\end{abstract}

\begin{keywords}{Keywords:}
Inception1D, Transformers, Parkinson's disease staging, H\&Y scale, VGRF Signals
\end{keywords}
\copyrightnotice

\section{Introduction}
\label{introduction}

The prevalence of Parkinson's disease is rising globally with an estimated 10 million people currently living with the condition \cite{Dorsey2007ProjectedNO}. This neurodegenerative disorder primarily affects the brain regions responsible for coordinating movement, leading to symptoms such as tremors and difficulties in walking. While there is no cure for Parkinson's disease, various treatments are available to control its symptoms. To better analyze the condition, gait analysis is commonly used to detect any irregularities in movement. This analysis is reliant on assessing a range of clinical symptoms and signs. These factors can pose challenges since they may overlap with other neurological disorders. For that, the extraction of gait information using foot sensors is essential to understand our movement patterns and identify abnormalities. To classify Parkinson’s disease into five distinct stages based on symptom severity, the Hoehn and Yahr (H\&Y) scale \cite{Goetz2004MovementDS} is commonly employed. Another frequently utilized method for assessing symptom severity is the Unified Parkinson’s Disease Rating Scale (UPDRS) \cite{Shulman2010TheCI}. Both of these scales play crucial roles in clinical settings and research for tracking disease progression and evaluating the effectiveness of treatments.

Despite advancements in gait-based diagnostic techniques, assessing the severity of the disease remains a significant challenge. Several deep learning-based techniques have been proposed for detecting Parkinson's disease (PD) from gait data,  and have produced encouraging results \cite{Narendra2021TheDO, Quan2021ADL, Erturul2016DetectionOP}. However, these methods are used for binary classification to detect PD based on gait patterns. The classification of PD severity into specific stages is less explored by the research community. Most of the methods use the H\&Y criteria derived from the publicly available Physionet gait dataset \cite{dataset}. Despite the good results obtained from the studies focusing on PD staging \cite{Naimi2023HCTHC, maachi}, they often face the problem of data imbalance, which causes a performance drop. This imbalance arises because certain severity stages, particularly the more advanced ones, are generally underrepresented, leading to biased learning and reduced classification performance. Models trained on such imbalanced data tend to overfit to the majority classes while struggling to correctly classify the minority classes, ultimately affecting their generalization capability.
 
In this work, we introduce \textit{InceptoFormer}, a Multi-Signal neural framework specifically designed for PD staging based on gait dynamics. Our approach not only improves classification accuracy but also ensures robust performance across imbalanced classes of gait data. By integrating multi-signal temporal feature extraction with attention mechanisms, \textit{InceptoFormer} effectively captures the complex temporal and spatial patterns in gait data, while employing an oversampling strategy to enhance the representation of minority classes. 

\section{Related Work}
\label{relatedwork}
Various deep-learning approaches have been applied to assess the severity stages of Parkinson's disease by analyzing gait data. In particular, the Vertical Ground Reaction Force (VGRF) signal is widely used, as it has been proven to be a crucial and distinguishing kinematic feature for detecting and evaluating Parkinson’s disease stages \cite{Guo2022DetectionAA}.
Ertugrul et al. \cite{Ertugrul} introduced an algorithm utilizing shifted 1D local binary patterns (1D-LBP) in conjunction with machine learning classifiers. Their approach involved applying a shifted 1D-LBP to construct 18 histograms of the corresponding patterns.
Zhao et al. \cite{Zhao} developed an algorithm featuring two parallel networks, a 2D Convolutional Neural Network (2D-ConvNet) to analyze the spatial distribution of forces, and a recurrent neural network (RNN) to examine temporal distributions. The final classification was determined by averaging the outputs from both networks.
Aşuroğlu et al. \cite{Tunç} explored a Perceiver-based multimodal model for predicting UPDRS scores from VGRF gait data. The Perceiver architecture leverages self-attention mechanisms to process sequences of varying lengths and complexities, significantly enhancing performance in severity assessment for Parkinson's disease.
Balaji et al. \cite{Balaji} proposed a correlation-based feature extraction method. Biomarkers were extracted from spatiotemporal VGRF gait data using correlation. They employed four supervised machine learning algorithms K-nearest neighbors (KNN), Naive Bayes (NB), Ensemble classifier (EC), and Support Vector Machine (SVM) to classify the severity of PD based on the Hoehn and Yahr (H\&Y) scale.
Naimi et al. \cite{Naimi1} introduced a hybrid ConvNet-Transformer model capable of capturing both spatial and temporal features. Their model is designed for both detection and severity evaluation of PD.
Veeraragavan et al.\cite{Veeraragavan} introduced an approach that uses feedforward neural networks (FNN) to classify severity levels.
Mirelman et al. \cite{Mirelman} employed random forest classifiers and support vector machines (SVM) to differentiate between stages of Parkinson's disease, identifying key features correlated with disease severity.

While existing methods have shown good performances in terms of classification accuracy for detecting and staging PD, they face two main challenges that hinder their overall performance: class imbalance and limitations in capturing the complex patterns of physiological data, mainly due to architectural constraints. These issues contribute to a noticeable drop in model performance. Our approach addresses these issues by integrating multi-signal feature extraction through Inception1D blocks to capture fine-grained temporal variations and by employing Transformer-based encoders to model both long-range dependencies and spatial correlations in gait dynamics. The proposed design mitigates class imbalance through a tailored data structuring and preprocessing strategy. It also enhances the ability to learn complex physiological patterns, ultimately leading to improved performance in PD severity evaluation.

\section{Method}
\label{method}

Our approach focuses on identifying the severity stage of Parkinson’s disease (PD) based on the Hoehn and Yahr (H\&Y) staging scale, which classifies symptoms into five stages. However, the Physionet gait dataset \cite{dataset}, used in most existing studies including ours, includes only four main stages: Stage 0 (healthy), Stage 1 (mild, severity 2), Stage 2 (moderate, severity 2.5), and Stage 3 (severe, severity 3). These stages are derived from gait signals collected through foot sensors attached to patients. Given the inherent complexity and variability of gait data, accurately distinguishing between these stages remains challenging, particularly in the presence of class imbalance due to the scarcity of data on advanced stages of the disease. The following sections provide a detailed presentation of our approach, including the data structuring and preprocessing strategy to deal with data imbalance and \textit{InceptoFormer} architecture for improving PD severity evaluation.

\subsection{Data Structuring and Preprocessing}

\begin{figure*}[b]
\centering
\begin{minipage}[b]{0.48\textwidth} 
\centering
\includegraphics[width=\textwidth]{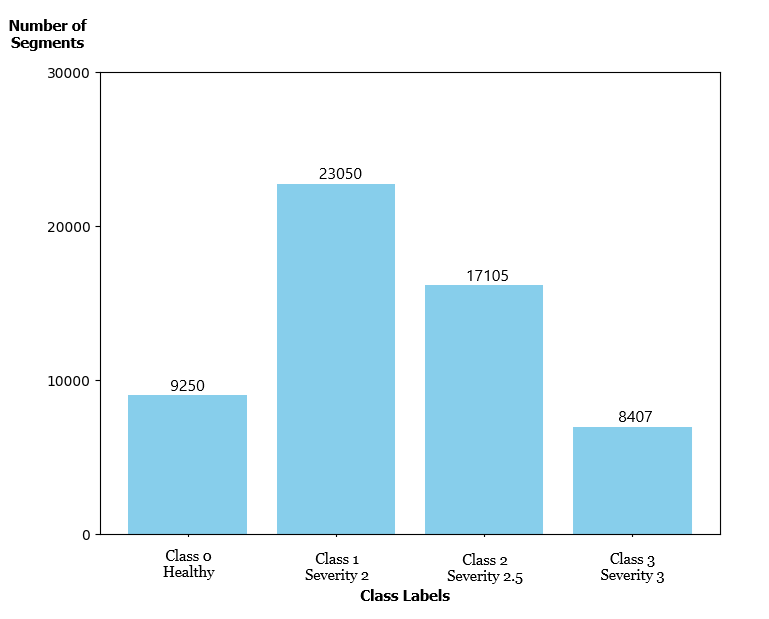}
\caption{Initial Class distribution: Before applying our data structuring and preprocessing strategy}
\label{fig:f322}
\end{minipage}
\hspace{0.02\textwidth} 
\begin{minipage}[b]{0.48\textwidth} 
\centering
\includegraphics[width=\textwidth]{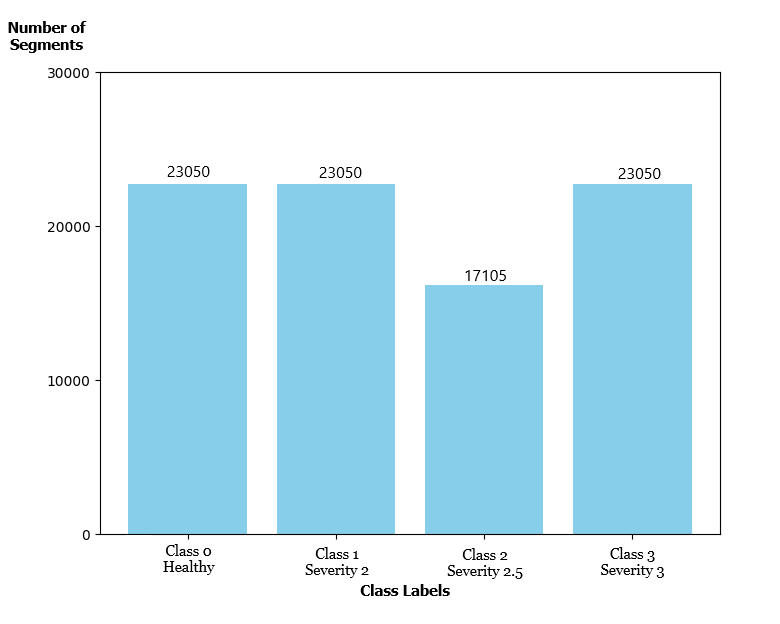}
\caption{Final Class distribution: After applying our data structuring and preprocessing strategy}
\label{fig:f319}
\end{minipage}
\end{figure*}

The Physionet dataset \cite{dataset} used in our study presents multiple 1D VGRF signals captured from patients' walks and measured using 18-foot sensors.
 We started by dividing the walks into small segments of 100-time steps with 50\% overlap composed of groups of elements. This segmentation not only increases the amount of training data but also preserves the temporal continuity of gait patterns, enhances feature extraction by capturing finer motion variations, and ensures a more robust representation of the gait dynamics across different severity levels.
 For each segment, we assign a specific label or category that identifies it. This is essential for time series data, as it helps capturing temporal patterns. 
 A common issue in the dataset used in our study is the imbalance between the four classes of the H\&Y scale as depicted in Figure  \ref{fig:f322}. For that, we employed an oversampling strategy of the minority class using the Synthetic Minority Over-sampling Technique \cite{Chawla2002SMOTESM} to balance the classes. 
 It is a preprocessing technique used to
address the class imbalance by creating synthetic samples. We
started by selecting the K-nearest neighbors of every sample that belongs to a minority class, then we generate samples along the line segments joining the minority class sample to its nearest
neighbors. The minority classes, class 0 and class 3 have been oversampled while majority
classes 1 and 2 maintain a stable number of samples, as they are sufficiently represented and
not considered minority classes.
Given the two minority class samples class 0 and class 3, respectively \( x_i \) and \( x_j \), the synthetic sample \( x_{\text{\textit{new}}} \) for each one is generated as follows:
\begin{equation}
x_{\text{\textit{new}}} = x_i + \lambda \cdot (x_j - x_i)
\end{equation}
where \( x_i \) and \( x_j \) are two randomly chosen samples from the minority class, and \( \lambda \) is a random number sampled from a uniform distribution: \( \lambda \sim U(0, 1) \).

For the minority classes 0 and 3, \( N_{c_i} \) the number of samples for class \( c_i \) and \( N_{\text{\textit{new}},\text{ }c_i} \) the number of synthetic samples generated for class \( c_i \), the synthetic samples are generated as follows:
\begin{equation}
N_{\text{\textit{new}},\text{ }c_i} = N_{\text{\textit{majority}}} - N_{c_i}, \quad c_i \in \{0, 3\}
\end{equation}
where \( N_{\text{\textit{majority}}} \) is the number of samples in the majority class class 1. The oversampling ensures that the number of samples in classes 0 and 3 equals that of the majority class:
\begin{equation}
N_{c_i} + N_{\text{\textit{new}},\text{ } c_i} = N_{\text{\textit{majority}}}, \quad c_i \in \{0, 3\}
\end{equation}
For classes 1 and 2, the sample count remains unchanged:
\begin{equation}
N_{\text{\textit{new}},\text{ }c_i} = 0, \quad c_i \in \{1, 2\}
\end{equation}

Figures \ref{fig:f322} and \ref{fig:f319} illustrate the class distribution before and after applying our data structuring and preprocessing strategy to the Physionet gait dataset.

\begin{figure}[b]
\centering
\includegraphics[width=0.9\textwidth]{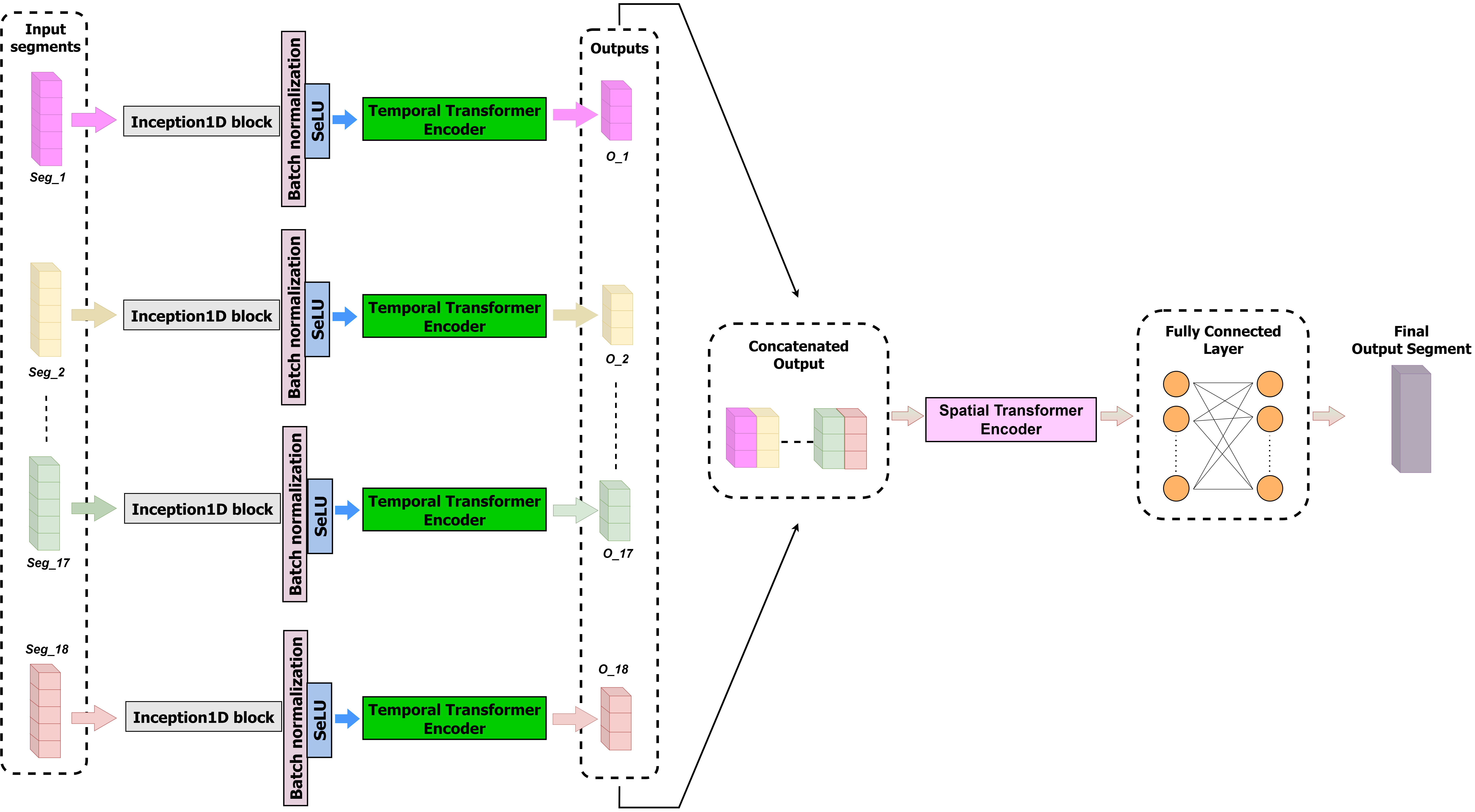}
\caption{The proposed InceptoFormer architecture. We have 18 signals captured from the sensors. To fully leverage the information in each signal, we process them independently using 18 parallel Inception1D blocks. The outputs are then concatenated and passed to a temporal transformer, and then we applied a spatial transformer encoder. In the end, a classifier block is used to generate the final classification.}
\label{fig:label}
\end{figure}

\subsection{InceptoFormer Architecture}

The detailed architecture of \textit{InceptoFormer} is provided in Figure \ref{fig:label}. This model is designed to predict the severity of Parkinson's disease based on gait analysis.
Since we are processing data from 18 sensors, we have 18 distinct signals, each capturing different aspects of the patient's foot movements during a walk. To fully explore the information in each signal,
we process them independently using 18 parallel Inception1D blocks to capture dependencies in different scales and extract unique features which are subsequently concatenated together to give a better description of the data. 
Following this stage, each feature stream is processed by a temporal transformer encoder, which models long-range dependencies within each gait sequence. This component enables the model to capture the progression of movement patterns over time, ensuring a more comprehensive understanding of gait dynamics. Before further processing, the outputs of these temporal transformers undergo dimensionality reduction to optimize computational efficiency.
The reduced feature representations from all 18 signals are subsequently concatenated to form a unified feature vector, which serves as the input to the spatial transformer encoder. It is responsible for capturing spatial dependencies between the sensors, learning how different regions of the foot interact during movement. By leveraging self-attention mechanisms, it effectively identifies correlations between sensor signals, refining the extracted feature representation.
The final stage consists of a classifier block used to predict the PD stage and generate the final classification. The following sections provide a more detailed breakdown of each \textit{InceptoFormer} component, highlighting their individual contributions to the overall architecture.

\begin{figure*}[b]
\centering
\includegraphics[width=1.0\textwidth]{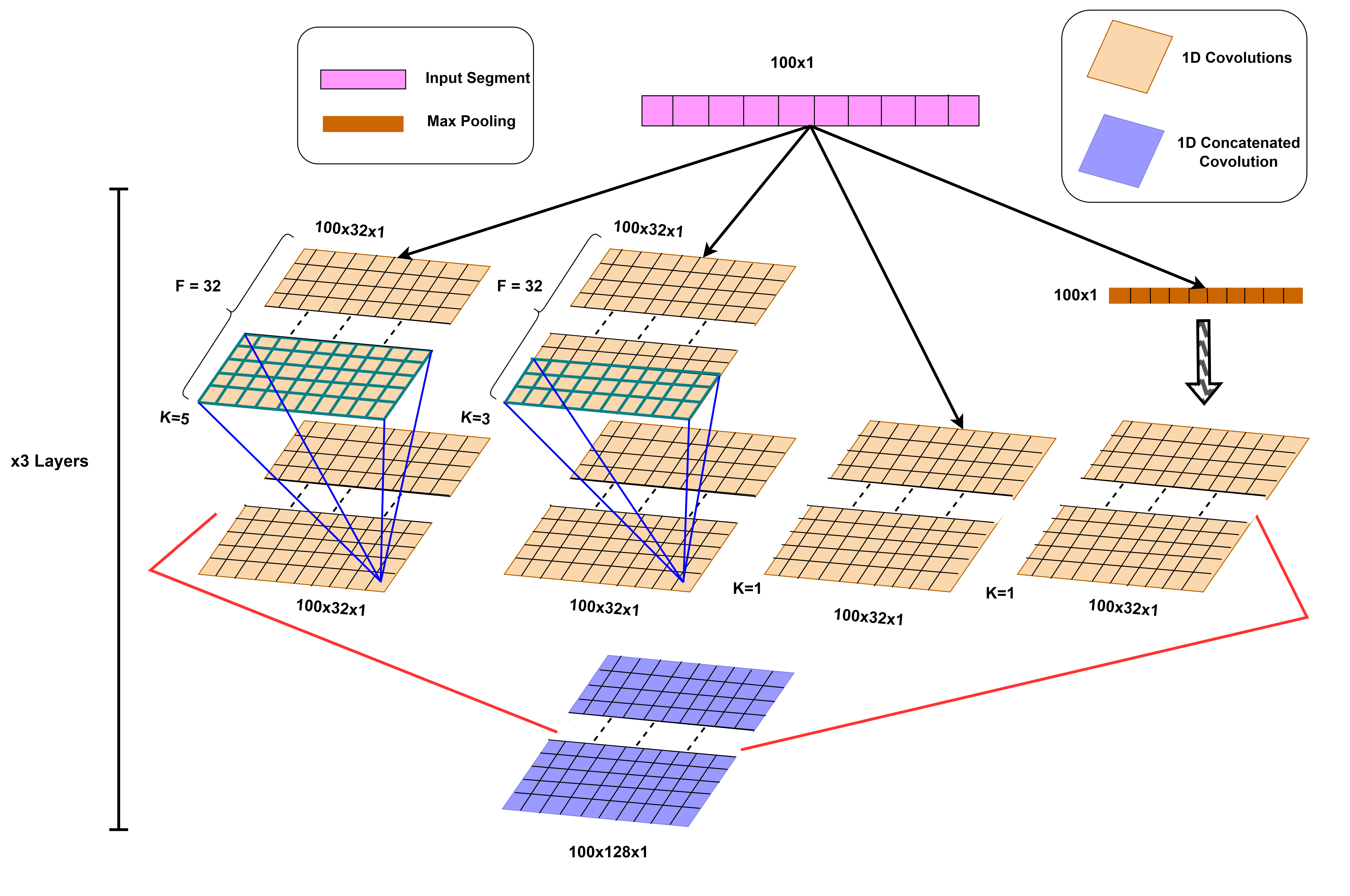}
\caption{Architecture of the Inception1D block}
\label{fig:incep1d}
\end{figure*}

\subsubsection{Inception1D Block}
We adapted the inception architecture which was initially designed for image processing for 1D data (see Figure \ref{fig:incep1d}). It introduces a new version of the inception module that can be easily integrated with our 1D VGRF signals.
It consists of three convolutional streams in parallel in which the previous layer is common to all of them. The first stream applies a 1D convolution with 32 filters and a kernel size of 1 (K=1). The second stream uses a convolutional layer with kernel size K=3, while the third stream employs a larger kernel size K=5, both having 32 filters as well. These streams allow for capturing features at various scales, with the smaller kernel sizes focusing on more localized patterns and the larger kernels extracting regional patterns. The convolutional layers in the Inception1D module can be summarized as follows:
\begin{equation}
y_k(t)=f\left(\sum_{i=0}^{k-1} W_k(i) \cdot x + b_k\right)
\end{equation}
with $x$ as the input segment, $i$ indexes the positions in the kernel, $W_k(i)$ represents the weights of the kernel of size $k$, $b_k$ is the bias term for the convolution with kernel size and $f(\cdot)$ is the SeLU activation function applied element-wise after convolution.

The 3 streams later converge to form a single output by concatenating their outputs.
This output is presented as:
\begin{equation}
y_{\text{\textit{inception}}} = [y_1, y_3, y_5]
\end{equation}
We follow the output with Batch Normalization to normalize the output from the module, improving the stability and speed of training, and an activation function is applied to introduce non-linearity. 
In this way, both local and regional information were extracted on the same features map. We repeated the inception architecture 3 times in cascade to achieve a deeper network, which made it possible to extract information from increasingly larger regions.

\subsubsection{Temporal Transformer Block} 

In \textit{InceptoFormer}, we implemented a temporal transformer encoder to minimize intraclass variance and capture long-range dependencies in the data. This encoder plays a key role in our architecture by encoding the input sequence and producing representations that reflect the underlying dependencies within the data \cite{Lohit2019TemporalTN}. This temporal transformer block is implemented after each Inception1D Block. It includes a multi-head attention layer with two heads, followed by a feed-forward network, mirroring the architecture used in BERT \cite{Devlin2019BERTPO} for natural language processing.

We applied a fixed positional encoding with a constant step size corresponding to the segment length since the gait data is segmented into fixed-length and constant intervals. The use of a fixed positional encoder enabled us to effectively capture the temporal patterns within these segments by maintaining the temporal ordering of the data. Furthermore, we applied normalization to the positional encoding to ensure the original vector information is not entirely masked.

\subsubsection{Spatial Transformer Block}

The outputs from the $18$ parallel temporal transformer encoders are concatenated. This step aids in creating a more compact representation of the data and eliminates redundant information. The resulting concatenated vector is then fed into the spatial transformer encoder block. A fixed positional encoding is also applied here, providing the spatial transformer encoder with details about the relative positions of elements within the concatenated vector, maintaining the spatial information of the sensors with respect to each other. Same as the temporal transformer encoder block, the spatial transformer includes a multi-head attention layer with two heads and a feed-forward network.

The primary function of the spatial transformer is to identify dependencies between the sensors by considering their spatial arrangement on the foot and potentially uncovering correlations among them.

\subsubsection{Classifier Block}

To predict the stage of Parkinson’s Disease, we employed a classifier consisting of two fully connected layers and an output layer, which serve as the final part of our \textit{InceptoFormer}. This block receives the output from the spatial transformer encoder and generates a probability distribution across the PD severity levels. The classifier was trained using the categorical cross-entropy loss function to adjust the weights and biases of the model. The resulting probabilities were then used to determine the predicted PD stage for each input signal.

\section{Experiments}

This section outlines the experiments conducted to evaluate the performance of our approach. We describe the dataset, the evaluation metrics, and the training details employed. Additionally, we discuss the results and illustrate the ablation study of our method.

\subsection{Dataset Description}

We used the Physionet gait dataset \cite{dataset} which contains multiple 1D VGRF signals recorded from patients' walks using 18 foot-mounted sensors. The dataset includes gait measurements from 93 patients with Parkinson’s disease (PD) and 73 healthy individuals. It was assembled by three research teams.
The first data was collected by Yogev et al. \cite{yogev}, it contains a gait cycle for normal walking on a leveled surface. The second data was reported by Hausdorff et al. \cite{hausdorff}, it contains the gait cycle for walking at a casual speed with RAS, and the final data was collected by Toledo et al. \cite{toledo}, it contains time series data of a subject for walking on a treadmill. This database includes
demographic information, measures of severity rating scale such as the Hoehn \& Yahr scale, UPDRS scale, and other related measures.

\subsection{Evaluation Metrics}

We evaluated our method using 10-fold cross-validation. We divided the Parkinsonian (Pd) and control (Co) groups into 10 folds, ensuring that each fold maintained the same dataset balance (70\% for Pd and 30\% for Co). For the evaluation metrics, we used the following notations: $TP$ for the true positives, $TN$ for the true negatives, $FP$ for the false positives, and $FN$ for the false negatives. Our method was assessed using precision, recall, F1-score, and accuracy. The utilized metrics equations are given below.
\begin{equation}
\text { \bf Precision: } Pr=\frac{TP}{TP+FP}
\end{equation}
\begin{equation}
\text { \bf Recall: } Re=\frac{TP}{TP+FN}
\end{equation}
\begin{equation}
\text { \bf F1-Score } =2\times \frac{ Pr \times Re}{ Pr+ Re}
\end{equation}
\begin{equation}
\text { \bf Accuracy (\%): } Acc =\frac{TP+TN}{TP+TN+FP+FN}\times 100\%
\end{equation}

\subsection{Training Details}

The proposed approach was trained and tested utilizing a batch size of 64 samples for each iteration. All fully connected layers, except for the output layer, employed the SeLU activation function. The model is trained using the Nadam stochastic optimization method \cite{nadam} which adjusts learning rates for each parameter, making it a robust optimizer for our approach. The learning rate is set to  
$\eta \ = 10^{\textbf{-}4}$.
We combine the Nadam optimizer with the Adam optimizer and Nesterov momentum. The update rule for the weights \( w \) in Nadam is given by:
\begin{equation}
w_{t+1} = w_t - \eta \cdot \left( \frac{\hat{m}_t}{\sqrt{\hat{v}_t} + \epsilon} \right)
\end{equation}
where \( \eta \) is the learning rate, \( \hat{m}_t \) is the moving average of the gradients, \( \hat{v}_t \) is the moving average of the squared gradients, and \( \epsilon \) is a small constant to avoid division by zero.
To enhance the performance of the \textit{InceptoFormer} and mitigate overfitting, we implemented a dropout rate of 0.2 and employed an early stopping based on the validation loss.

\subsection{Results}

\begin{table*}[b]
\centering
\caption{Comparison of our proposed architecture with the state-of-the-art methods for PD severity evaluation. Method marked with * indicates that we retrained it using the SMOTE technique and the same settings as reported in the corresponding paper. The best results are in bold and the second best results are in italic red.}
\begin{tabular}{|l|c|c|c|c|}
\hline
 Method & Accuracy & Precision & Recall & F1-Score \\ \hline
 OF-DDNet \cite{lu} & 84.00\% & \textcolor{red}{\textit{90.25}\%} & 86.00\% & --- \\ \hline
 1D-ConvNet \cite{maachi} & 85.30\% & 89,48\% & 82.68\% & 85.04\% \\ \hline
Feed Forward Network \cite{Veeraragavan} & 86.50\% & 87.73\% & 87.55\% & 87.67\% \\ \hline
1D-Convolutional Transformer \cite{Naimi1} & 88.00\% & 87.25\% & 85.25\% & 85.50\% \\ \hline
1D-Convolutional Transformer * \cite{Naimi1} & \textcolor{red}{\textit{89.12}\%} & 89.52\% & \textcolor{red}{\textit{89.25}\%} & \textcolor{red}{\textit{88.11}\%} \\ \hline
 \textbf{InceptoFormer (Ours) }& \textbf{96.60\%} & \textbf{93.97\%} & \textbf{94.15\%} & \textbf{93.60\%} \\  \hline
\end{tabular}
\label{tab:example1000}
\end{table*}

Table \ref{tab:example1000} presents a comparative analysis between our proposed approach and several existing methods. Through an examination of the average metric variations across these methods, we show that \textit{InceptoFormer} outperforms the others in terms of precision, recall, and F1-score, achieving a superior final accuracy of 96.6\%. This performance is largely attributed to the integration of Inception1D, which enables the model to capture critical multi-scale temporal features, and the Transformer-based design, which effectively models long-term dependencies and intricate gait dynamics. Furthermore, a key factor contributing to this performance gain is our data structuring and preprocessing strategy. By addressing the severe class imbalance inherent in the Physionet dataset, we ensure a better representation of all severity levels, preventing the model from being biased toward the majority classes. Table \ref{tab:example1000} demonstrates the effectiveness of our data structuring and preprocessing strategy. We retrained the method marked with * using this strategy under the same settings as reported in its corresponding paper. Notably, the application of our strategy led to a measurable performance improvement as seen in the enhanced accuracy, precision, recall, and F1-score. This further highlights the necessity of addressing the class imbalance in PD severity evaluation and underscores the robustness of our proposed approach in PD staging compared to other methods on the same Physionet dataset.

\begin{figure}[t]
     \centering
		\includegraphics[scale=0.5]{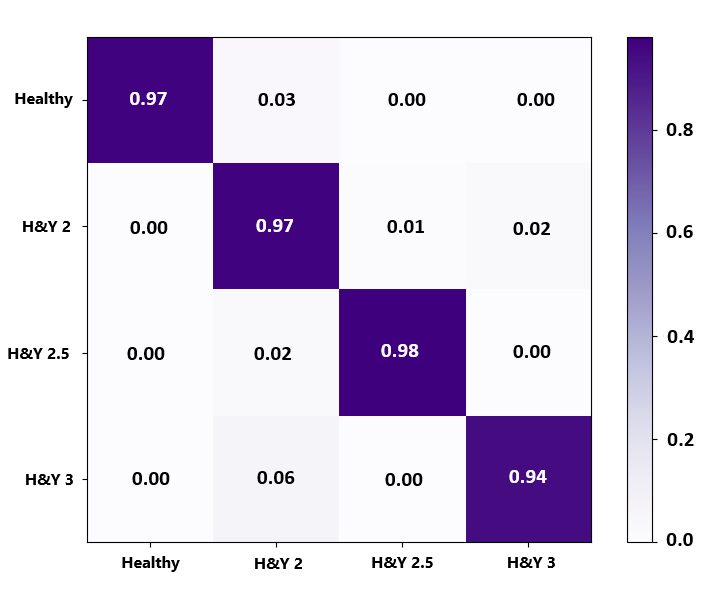}
	\caption{Confusion matrix of InceptoFormer}
	\label{FIG:102}
\end{figure}

Figure \ref{FIG:102} illustrates the average confusion matrix across all 10 folds for our proposed approach. We achieved a 97\% accuracy in classifying instances of Healthy patients. For severity 2 cases, 97\% were accurately predicted, with minor misclassifications into severity 2.5 and severity 3. The model demonstrated good performance in classifying severity 2.5 cases, correctly predicting 98\%, with a slight tendency to misclassify them as severity 2. In the case of severity 3 patients, the model accurately classified 94\% of instances. However, this category experienced a notable degree of misclassification into severity 2.

\subsection{Ablation Study}
To evaluate the contribution of each key component in \textit{InceptoFormer}, we conducted ablation experiments on the Physionet gait dataset to assess their impact on the final performance in terms of accuracy. A detailed analysis is provided in Table \ref{tab:example100}.
In Model 1, we removed both the Temporal and Spatial Transformer blocks, retaining only the Inception1D module. This configuration resulted in a significant drop in performance, with accuracy decreasing by 18.26\%, precision by 13.82\%, recall by 13.33\%, and the F1-Score by 17.83\%. These results underscore the critical role played by the Temporal and Spatial Transformer blocks in capturing the temporal and spatial dependencies that are essential for recognizing gait patterns. The absence of these attention mechanisms leaves the model reliant solely on multi-scale feature extraction from the Inception1D component, which, while useful, is insufficient to capture complex temporal relationships, leading to reduced overall performance.
Conversely, in Model 2, we excluded the Inception1D module and retained the Temporal and Spatial Transformer blocks. The model performance also degraded but to a lesser extent than in Model 1, with accuracy decreasing by 7.43\%, precision by 6.88\%, recall by 1.27\%, and the F1-Score by 5.98\%. These results highlight the importance of the Inception1D component, which enhances the model capacity to extract multi-scale features from the input data, thereby improving the model ability to differentiate between subtle variations in gait patterns. However, the Transformer blocks were able to compensate for the absence of multi-scale features to some degree, emphasizing their strong contribution to the overall effectiveness of our model.
Finally, Model 3, representing our \textit{InceptoFormer} that integrates both the Temporal and Spatial Transformer blocks alongside the Inception1D module, achieves the best performance across all metrics. The combination of these components leads to the highest accuracy (96.6\%), precision (93.97\%), recall (94.15\%), and F1-Score (93.6\%), demonstrating that the synergy between multi-scale feature extraction and attention-based temporal-spatial modeling is essential for superior performance in gait analysis tasks. This comprehensive integration allows our \textit{InceptoFormer} to capture both local and global dependencies, offering a more nuanced and complete understanding of gait dynamics.

\begin{table*}[t]
\centering
\small
\caption{Ablation study results on the effect of each component of our InceptoFormer.}
\begin{tabular}{|c|c|c|c|c|c|c|}
\hline
 Variations & Inception1D & \makecell{Temporal and \\ Spatial Transformers} & Accuracy (\%) & Precision (\%) & Recall (\%) & F1-Score (\%) \\ \hline
 Model 1 & \ding{51} & \ding{55} & 78,34 (↓18.26) & 80.15 (↓13.82) & 80.82 (↓13.33) & 75.77 (↓17.83) \\ \hline
 Model 2  &  \ding{55} & \ding{51} &  89.17 (↓7.43) & 87.09 (↓6.88) & 92.88 (↓1.27) & 87.62 (↓5.98) \\ \hline
 Model 3 & \ding{51} & \ding{51} &  \textbf{96.6} & \textbf{93.97}& \textbf{94.15} & \textbf{93.6} \\   \hline
\end{tabular}
\label{tab:example100}
\end{table*}

\section{Conclusions}
\label{sec:conclusion}
In this work, we introduced a multi-signal neural framework for Parkinson’s disease severity evaluation using gait dynamics, integrating Inception1D for multi-scale feature extraction and transformer encoders for temporal and spatial modeling. We proposed a data structuring and preprocessing strategy to address class imbalance and improve classification robustness. Our model achieved 96.6\% accuracy on the Physionet dataset, surpassing existing methods. Our findings highlight the potential of our framework for fine-grained Parkinson’s disease staging, which could be expanded to other clinical applications.



\printbibliography[heading=subbibintoc]

\end{document}